\pdfoutput=1

\documentclass[11pt]{article}

\usepackage[review]{acl}

\usepackage{times}
\usepackage{latexsym}

\usepackage[T1]{fontenc}

\usepackage[utf8]{inputenc}

\usepackage{microtype}

\usepackage{inconsolata}

\usepackage{graphicx}

\title{Team Ryu's Submission to SIGMORPHON 2024 Shared Task on Subword Tokenization}

\author{Zilong Li \\
  University of Colorado, Boulder\\
  \texttt{zili1126@colorado.edu} \\}

\begin{document}
\maketitle
\begin{abstract}

This papers presents the submission of team Ryu to the canceled SIGMORPHON 2024 shared task on subword tokenization. My submission explores whether morphological segmentation methods can be used as a part of subword tokenizers. I adopt two approaches: the statistical segmentation method Morfessor and a transformer based sequence-to-sequence (seq2seq) segmentation model in tokenizers. The prediction results show that morphological segmentation could be as effective as commonly used subword tokenizers. Additionally, I investigate how a tokenizer's vocabulary influences the performance of language models. A tokenizer with a balanced token frequency distribution tends to work better. A balanced token vocabulary can be achieved by keeping frequent words as unique tokens.

\end{abstract}

\section{Introduction}

Subword tokenization has become a widely accepted step in the pipeline of many NLP applications. People adopt subword tokenization, because it can keep frequently used words as they are, break long words into short pieces to reduce the size of vocabulary, and handle out-of-vocabulary (OOV) words. \citet{wolleb-etal-2023-assessing} summarized advantages of subword tokenization into three key aspects: \textbf{Frequency}, \textbf{Compositionality} and \textbf{Unknown words}. Although subword tokenization has been proved effective, few people have explored which part of advantages plays the most important role. Additionally, it is also difficult to evaluate different methods of tokenization without investigating their performance on downstream tasks.

Recently, transformer-based large language models (LLMs) have established their unchallengeable status in every field of NLP. It is widely accepted that a language model with a more parameters trained on a vast mount of data will perform better. However, compared to LLMs which require billions of word tokens to learn languages, children who are exposed to fewer than 100 million word tokens by age 13 are more efficient language learners \citep{children-language-acquisition}. Here people come across the question, whether we can build a more data efficient language model, so that it can achieve relatively good performance with a smaller data consumption. With a data efficient LM, it is easier for people to explore the influence of subword tokenization, as a small size of training data is more sensitive to the method of tokenization.

This paper describes my submission to the SIGMORPHON 2024 Subword Tokenization shared task. Participants are asked to develop a subword tokenization system and use it to pretrain a language model on the 100M word tokens dataset from the BabyLM challenge \citep{warstadt-etal-2023-findings}. The performance of pretrained model is evaluated by model's predictions after fine-tuning it on three different subtasks: \textbf{Word and Definition}, \textbf{Word and Morphology} and \textbf{Word and Word}. In this paper, I introduce two subword tokenization systems with their variants: one based on a statistics-based morphological segmentation method and the other based on a neural seq2seq model. This paper basically describes my systems and seeks to verify conclusions of other studies based on my discoveries.

\section{Related work}

\subsection{Subword Tokenization}

Although currently frequently used subword tokenization methods such as Byte-Pair Encoding (BPE), WordPiece, Unigram and SentencePiece have proven effective through the success of various LLMs, the source of their effectiveness is still unclear. \citet{mager-etal-2022-bpe} compared the performance of different morphological segmentation methods with the BPE in four low-resource languages on machine translation and morphological segmentation tasks. They tested both unsupervised methods and supervised methods. They found that both of them outperform the BPE in the task of morphological segmentation, but the latter group work much better. However, in the machine translation task, unsupervised methods outperform supervised ones by a large margin and the BPE ranks very high in many experiments. \citet{wolleb-etal-2023-assessing} developed a special tokenization method based on the Huffman tree. In their method, the most frequent words are represented with a single symbol, while less frequent words are a composition of symbols. This tokenization only possesses the advantage of frequency, because symbols in the Huffman tree are not semantically related. They discovered that with more independent symbols, the Huffman tree tokenizer can achieve up to 90\% power of BPE in machine translation. It is concluded that the strength of BPE mainly comes from its ability to keep frequent words as unique tokens. \citet{zouhar-etal-2023-tokenization} proposed a metric also based on the knowledge of information theory. Rényi entropy here is introduced to measure a tokenization's distribution. They found a strong correlation between models' performance in machine translation and the score of Rényi entropy score. Their experimental results reveal that an unbalanced token distribution, that is, a tokenization with very high frequency tokens, is harmful to models' performance.

\subsection{BabyLM}

The first BabyLM challenge in 2023 asked participants to submit data efficient models pretrained on 10M or 100M word tokens data. 31 teams participated in this challenge. Model \textbf{ELC-BERT} \citep{georges-gabriel-charpentier-samuel-2023-layers} is the winner in both 10M and 100M tracks. ELC-BERT is a variant of the BERT base model, different from the BERT base model in the position of normalization layer, activation function, attention mechanism, feed-forward weight initialization and layer weighting. ELC-BERT stands out from 162 model submissions and demonstrates its capability, which is comparable to Llama2 and RoBERTa-Base. The success of ELC-BERT shows that we can still enhance capabilities of language models by modifying their architecture.

Although there are totally 162 models submitted to this challenge, few participants discuss the influence of tokenization method on the performance of models. Most groups adopt commonly used tokenizers such as BPE, WordPiece and Unigram. Groups who take tokenization into consideration primarily explore the influence of vocabulary size. \citet{jumelet-etal-2023-chapgtp} is the only group trying to introduce other method of tokenization. They make use of the FLOAT tokenizer \citep{hofmann-etal-2022-embarrassingly} to build a vocabulary which adheres to the morphological formation of English words. Their result reveals that the FLOAT tokenizer cannot promote the language model and even decreases its score on downstream tasks by 20 percents.

\section{Data and Task Description}

The training data for the pretrained language model is the same as the data used in the BabyLM challenge's strict track. It contains 98.04M word tokens extracted from dialogue, children's books, subtitles, Wikipedia and other sources. Additional data, which is used to train the tokenizer, comes from the SIGMORPHON 2022 Morpheme Segmentation shared task \citep{batsuren-etal-2022-sigmorphon}. Only English morphological segmentation data is used, and it is in the form of word-segmentation pairs.

The pretrained model is evaluated on three subtasks. Each subtask has its own training, validation and test data provided by the organizer in the form of TSV file.

\paragraph{Word and Definition} is the task to classify whether a given word and its definition match with each other. Each line in the data file consists a word and its definition as well as a label. Label 0 indicates that this word doesn't match the definition, while 1 means that the definition is correct.

\paragraph{Word and Morphology} asks language model to classify whether a given word contains a specified morphology. Each line is a tuple of the target word, its possible formation and a label. Label 0 means that the target word contains the given morphology, while 1 denotes that the target word does not match the morphology.

\paragraph{Word and Word} is also a classification task to judge whether two given words have a semantic relationship. Each line contains two words and a label. Label 1 indicates that these two words are semantically related. Label 0 means that there is no obvious relationship between these two words.

\section{System Design}

\subsection{Motivation}

As the advantage of subword tokenization can be concluded as frequency, compositionality and unknown words, we can build several tokenizers to investigate how each part of advantages plays its role. It is believed that a tokenizer that aligns more closely with morphological formation will produce better results. Therefore, morphological segmentation methods are incorporated into tokenizers. Because the frequency of tokens is directly related to the vocabulary, different vocabularies are applied to the same tokenizer to create its variants.

\subsection{Data Preprocessing}

The raw data of the BabyLM challenge is dirty and not suitable for training. To get rid of punctuation marks, website URLs and characters from other languages, preprocessing codes from ELC-BERT were used. After preprocessing, I counted the frequency of each word and sorted them in descending order. There are 500k distinct words appearing in the preprocessed training data. Since words with low frequency may come from other languages or may be misspellings, only words appearing more than 30 times are considered. Finally, 53k words along with their frequency are used for tokenization.

\subsection{Morfessor Based Tokenizer and its Variants}

Morfessor \citep{morfessor_1} is a probabilistic machine learning method used to segment words into possible morphs from raw text data. It segments words by maximizing the product of the model's prior probability and the data likelihood. The prior probability includes an estimated morph length probability distribution and a morph frequency probability distribution. The data structure of splitting trees is built and maintained with each node representing a possible morph. Morfessor 2.0 \citep{smit-etal-2014-morfessor} is an extension of the baseline model. It implements a semi-supervised method so that annotated data can be used to guide the formation of morphs.

\paragraph{Type Based Tokenizer} is derived from the morfessor-baseline model. 53k distinct training words are directly put into the morfessor model. Words' frequency is not considered this time, so the model segments words by their types and results in a relatively small vocabulary. To semi-supervise the segmentation model and also reduce the influence of data distribution in SIGMORPHON 2022 Morpheme Segmentation shared task, I sample only 3k words and use their true segmentation as the annotated input to the model. The sampling is base on the word frequency. Because a long word has a higher probability to be a combination of morphemes, word length is used to modify the frequency and form the corresponding sampling probability:

\begin{equation}
  \label{sampling_score}
  p_w = \frac{l_w f_w}{\sum_{i = 1}^{k} l_i f_i}
\end{equation}

After completing the training, the whole annotated data from SIGMORPHON 2022 will be used as the development data to optimize the model's likelihoods weights.

To differentiate tokens that appear at the beginning of words from those that appear later in words, I introduce a special symbol Ġ to the first token of each word, so word \textit{unsupervised} would be tokenized as:

\begin{center}
    Ġun \quad super \quad vis \quad ed
\end{center}

All words appearing more than 30 times will be segmented again by the model to build a vocabulary. Besides tokens from words, special tokens, punctuation marks and special symbols are added to the vocabulary.

When tokenizing a word, the morfessor model within the toknizer will firstly be called. For each fragment the after segmentation, if it exists in the vocabulary, it will be retained. If not, it will be replaced by the special token \textbf{[UNK]}.

\paragraph{Token Based Tokenizer} is different from the previous tokenizer in that it takes words' frequency in account when building splitting trees. A \textit{log} function is used to compress the frequency so that the morfessor model isn't influenced by the extremely skewed data distribution and still behaves well in morph segmentation.

\begin{equation}
  \label{log_function}
  f_{new} = log_{30}(f_{old} + 1)
\end{equation}

Apart from words' frequency and the reshape function, the rest of token based tokenizer is the same as the type based tokenizer above.

\paragraph{Tokenizer with Frequent Words} keeps the most frequent words intact, but applies the same segmentation method to less frequent words as the type based tokenizer above. The threshold to determine whether a word is frequent or not is 1700, so finally around 4100 words are retained and the tokenizer has a relatively balanced vocabulary.

\subsection{Neural Based Tokenizer}

Neural networks have been widely applied to the task of morpheme segmentation and achieved good results \citep{batsuren-etal-2022-sigmorphon}. In this paper, I construct a transformer based seq2seq model trained on SIGMORPHON 2022 English word segmentation data as a component of the tokenizer. This neural model has 6 encoder and decoder transformer layers. A Unigram tokenizer is employed to preprocess the data and constrain the output. Implementation details and hyperparameters are provided in table \ref{tab:neural_tokenizer}.

\begin{table}[h]
  \centering
  \begin{tabular}{lc}
    \hline
    \textbf{Hyperparameters} & \textbf{Values} \\
    \hline
    \verb|Vocabulary|    & 8000           \\
    \verb|Layers|    & 6           \\
    \verb|Attention head|     & 8           \\
    \verb|Embedding size|     & 256           \\
    \verb|Feedfoward dim|     & 1024           \\
    \verb|Dropout|     & 0.3           \\
    \verb|Batch size|      & 4096            \\
    \verb|Learning rate|     & 0.001           \\
    \verb|Epochs|     & 400           \\
    \hline
  \end{tabular}
  \caption{Hyperparameters of neural segmentation model}
  \label{tab:neural_tokenizer}
\end{table}

The vocabulary building and tokenization processes are the same as tokenizers described above. Only words with frequency more than 30 are used to build the vocabulary. For a sequence of fragments tokenized from a word, the special symbol Ġ will be added to the first fragment.

\section{Experimental Details}

\subsection{Pretraining}

To compare the performance of different tokenizers, I select the ELC-BERT model's structure and its basic pretraining strategies. The models are pretrained on the task of predicting a masked token in a sequence. Among the 15\% masked tokens, 80\% of them are masked as \textbf{[MASK]}, while 10\% of them are left unchanged, and the remaining 10\% are replaced randomly by other tokens. For the first 90\% of the pretraining steps, the sequence length is set to 128. After this, it is increased to 512. Because of the budget limitation, I adopt hyperparaters proposed by \citet{izsak-etal-2021-train}. Model structure and hyperparameters in detail are shown in table \ref{tab:training}.

\begin{table}[h]
  \centering
  \begin{tabular}{lc}
    \hline
    \textbf{Hyperparameters} & \textbf{Values} \\
    \hline
    \verb|Layers|    & 12           \\
    \verb|Attention head|     & 12           \\
    \verb|Embedding size|     & 768           \\
    \verb|Feedfoward dim|     & 2048           \\
    \verb|Dropout|     & 0.1           \\
    \verb|Batch size|      & 4096            \\
    \verb|Learning rate|     & 0.001           \\
    \verb|Steps|     & 31250           \\
    \verb|Warmup proportion|     & 0.02           \\
    \hline
  \end{tabular}
  \caption{Hyperparameters of pretrained language models}
  \label{tab:training}
\end{table}

\subsection{Fine-tuning}

After pretraining, each model is fine-tuned on three subtasks. For each input pair, a classification layer is added on the top of the \textbf{[CLS]} token. Due to the relatively small number of parameters in pretrained models, full-parameter fine-tuning method is used. I set an early stopping point when the loss value doesn't go down for 5 consecutive epochs. For evaluation, I pick out the best model for each subtask independently. Table \ref{tab:finetuning} demonstrates the fine-tuning details.

\begin{table}[h]
  \centering
  \begin{tabular}{lc}
    \hline
    \textbf{Hyperparameters} & \textbf{Values} \\
    \hline
    \verb|Epochs|      & 10            \\
    \verb|Early stopping epoch|      & 5            \\
    \verb|Batch size|      & 32            \\
    \verb|Learning rate|     & 5e-5           \\
    \verb|Dropout|     & 0.5           \\
    \hline
  \end{tabular}
  \caption{Fine-tuning setup}
  \label{tab:finetuning}
\end{table}

\section{Result and Analysis}

\subsection{Prediction Result}

The prediction accuracy for each model on three subtasks is shown in the table below.

\begin{table}[h]
  \centering
  \begin{tabular}{lccc}
    \hline
    \textbf{Model} & \textbf{WaD} & \textbf{WaM} & \textbf{WaW} \\
    \hline
    \verb|Morfessor Type|  & 0.718  & \textbf{0.872}  & 0.736 \\
    \verb|Morfessor Token|  & 0.736  & 0.867  & 0.755 \\
    \verb|Morfessor Frequency|  & 0.743  & 0.869  & 0.759 \\
    \verb|Neural|  & \textbf{0.75}  & 0.856  & 0.727 \\
    \verb|ELC BERT|  & 0.715  & 0.858  & \textbf{0.773} \\
    \hline
  \end{tabular}
  \caption{Prediction accuracy}
  \label{tab:accuracy}
\end{table}

We can learn from the table that the neural segmentation model based tokenizer achieves the best results in the Word and Definition task, but it is not the best one in the other two subtasks. All tokenizers developed in this paper perform equivalently or better than the WordPiece tokenizer in the Word and Morphology subtask. However, in the Word and Word subtask, the WordPiece tokenizer outperforms all other tokenizers.

Among three morfessor based tokenizers, the one with the most frequent words retained as independent tokens performs better on average. The word type morfessor tokenizer reaches the highest score in the morphological task, but it performs less effectively in the other two tasks.

\subsection{Analysis}

Although the token based morfessor tokenizer differs from the type based tokenizer in the algorithm, we can still compare them by examining their vocabulary distribution. Token based morfessor tokenizer has 6416 tokens that don't show in the vocabulary of type based tokenizer. Among these tokens, 4400 tokens are independent words that are not frequent (that is, each token appears less than 1700 times in the training data). Therefore, token based morfessor tokenizer can be regarded as a variant of type based tokenizer with a part of infrequent words kept as independent tokens.

\begin{figure}[h]
  \includegraphics[width=\columnwidth]{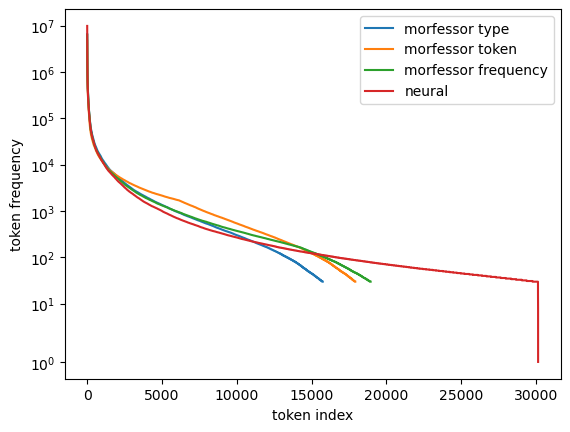}
  \caption{Token frequency distribution of tokenizers}
  \label{fig:distribution}
\end{figure}

According to the work of \citet{wolleb-etal-2023-assessing} and \citet{zouhar-etal-2023-tokenization}, the distribution of tokens may be related to the effectiveness of tokenizers. Figure \ref{fig:distribution} demonstrates the token frequency distribution for four different tokenizers. The neural based tokenizer has the largest vocabulary size, but it has a relatively unbalanced distribution, because its vocabulary contains more tokens with very low frequency. The other three morfessor based tokenizers have similar vocabulary. The token based tokenizer differs from the type based one in that it has more tokens with frequency between 100 and 1000, while the morfessor tokenizer which keeps the most frequent words as independent tokens has more tokens that appear between 1000 and 10000 times.

\begin{table}[h]
  \centering
  \begin{tabular}{lc}
    \hline
    \textbf{Model} & \textbf{Entropy} \\
    \hline
    \verb|Morfessor Type|  & 9.288 \\
    \verb|Morfessor Token|  & 9.298 \\
    \verb|Morfessor Frequency|  & 10.023 \\
    \verb|Neural|  & 9.136 \\
    \hline
  \end{tabular}
  \caption{Shannon Entropy of tokenizers}
  \label{tab:entropy}
\end{table}

Shannon entropy can be used to measure the discreteness of a distribution. A distribution with a higher entropy value tends to be more evenly distributed. Table \ref{tab:entropy} presents entropy value for each tokenizer's vocabulary. The tokenizer which keeps the most frequent words as tokens has the highest entropy value, while the neural model based tokenizer has the lowest entropy value. 

Combining the prediction results and the vocabulary distributions of tokenizers, we may observe a weak correlation between the performance of tokenizers and their tokens' frequency distribution. For tokenizers with similar vocabulary sizes, the one with a more balanced vocabulary tends to perform better than others. An easy way to make a tokenizer more balanced is to keep words as unique tokens. Compared to infrequent words, keeping the most frequent words may have a greater promotion of model's performance. Currently used subword tokenization methods, such BPE and WordPiece, also tend to retain the most frequent words as tokens.

According to the prediction results, morphological segmentation based tokenizers could perform equivalently to commonly used tokenization methods. In the morphological task, morphological segmentation based tokenizers commonly outperform currently used tokenizers, like Wordpiece in the ELC-BERT model. This is likely because they segment words into tokens that align more closely with morphemes in natural languages. Although neural models perform perfectly in the morpheme segmentation task, they are not suitable for tokenization, due to the issue of neural model's hallucination. Neural model may generate morphs that don't exist in natural languages. Even though these nonexistent morphs couldn't influence classification tasks, they may expand the vocabulary size and make its frequency distribution more unbalanced. Consequently, many meaningless tokens will be learned by language models, and this will weaken models' ability of understanding.

\section{Conclusion}

This paper describes my submission to the SIGMORPHON 2024 shared task on subword tokenization. In this paper, I build tokenizers using two commonly used morphological segmentation methods, morfessor and transformer based neural model. Another two variants of morfessor tokenizer are created to investiagte the influence of vocabulary.

The results show that morphological segmentation based tokenizers could work as well as commonly used tokenizers. Vocabulary has a small impact on the model's performance. Generally, a tokenizer with a more balanced token frequency distribution behaves better. An easy way to implement this is to keep the most frequent words as unique tokens. However, how to find an optimal point at which tokenizers can deal with OOV words but still have a relatively balanced vocabulary is still an open question. This question needs to be addressed in the future work.

\bibliography{custom}

\end{document}